\documentclass[10pt]{article}
\usepackage[a4paper,left=22mm,right=22mm,top=30mm,bottom=28mm,headheight=30pt,headsep=20pt,footskip=30pt]{geometry}
\usepackage{fontspec}
\usepackage{fix-cm}
\usepackage{amsmath,amssymb,graphicx,booktabs,tabularx,array,xcolor}
\usepackage[unicode,colorlinks=true,linkcolor=blue!45!black,citecolor=blue!45!black,urlcolor=blue!45!black]{hyperref}
\IfFileExists{/System/Library/Fonts/Supplemental/Charter.ttc}{
\setmainfont{Charter.ttc}[Path=/System/Library/Fonts/Supplemental/,FontIndex=0,BoldFont=Charter.ttc,BoldFeatures={FontIndex=3},ItalicFont=Charter.ttc,ItalicFeatures={FontIndex=1},BoldItalicFont=Charter.ttc,BoldItalicFeatures={FontIndex=2}]
}{\IfFileExists{XCharter-Regular.otf}{
\setmainfont{XCharter-Regular.otf}[BoldFont=XCharter-Bold.otf,ItalicFont=XCharter-Italic.otf,BoldItalicFont=XCharter-BoldItalic.otf]
}{
\setmainfont{texgyrepagella-regular.otf}[BoldFont=texgyrepagella-bold.otf,ItalicFont=texgyrepagella-italic.otf,BoldItalicFont=texgyrepagella-bolditalic.otf]
}}
\usepackage{tikz}
\makeatletter

\newcommand{\inlinecaption}[3]{\begingroup\def\@captype{#1}\caption{#2}\label{#3}\endgroup}
\newcommand{\paperheadtitle}{SkillDuet: Co-Evolving Models and Skills through Response Feedback}
\newcommand{\paperheadrule}{{\color{black!55}\hrule height .4pt width \textwidth}}
\def\ps@paper{%
  \def\@oddhead{\vbox to\headheight{\vfil\hbox to\textwidth{\hfil\normalfont\fontsize{8}{10}\selectfont\paperheadtitle\hfil}\kern4pt\paperheadrule}}%
  \let\@evenhead\@oddhead
  \def\@oddfoot{\vbox{\paperheadrule\kern5pt\hbox to\textwidth{\hfil\normalfont\fontsize{8}{10}\selectfont\thepage}}}%
  \let\@evenfoot\@oddfoot
}
\def\ps@brandfirst{%
  \def\@oddhead{\vbox to\headheight{\vfil\hbox to\textwidth{\includegraphics[height=6mm]{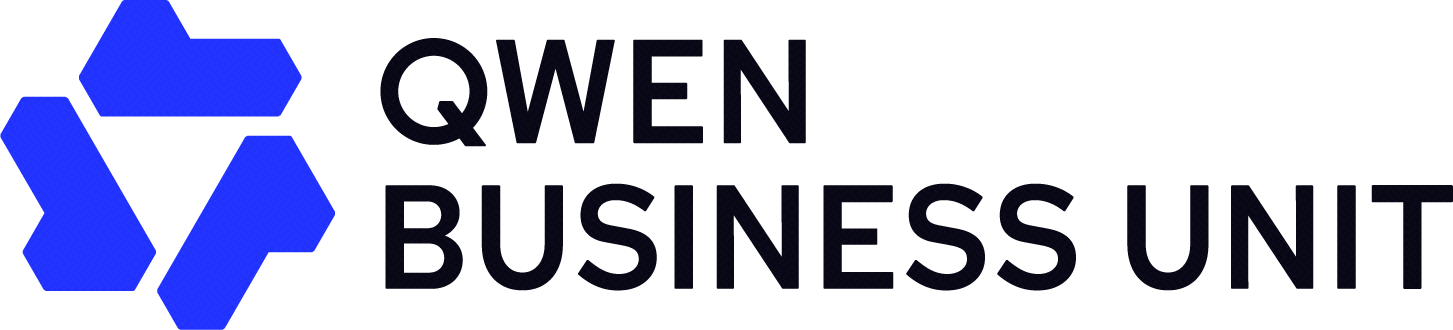}\hfil}\kern9pt\paperheadrule}}%
  \let\@evenhead\@oddhead
  \def\@oddfoot{\vbox{\paperheadrule\kern5pt\hbox to\textwidth{\hfil\normalfont\fontsize{8}{10}\selectfont\phantom{\thepage}}}}%
  \let\@evenfoot\@oddfoot
}
\AtBeginDocument{\pagestyle{paper}}
\renewcommand{\maketitle}{\thispagestyle{brandfirst}\@maketitle\suppressfloats[t]}
\renewcommand{\@maketitle}{%
  \begingroup
  \setlength{\parindent}{0pt}\setlength{\parskip}{0pt}
  {\raggedright\fontsize{21}{23}\selectfont\bfseries\@title\par}
  \vspace{16pt}
  {\raggedright\fontsize{9}{12}\selectfont\bfseries\@author\par}
  \endgroup
  \par\vspace{13pt}
}
\renewcommand{\@seccntformat}[1]{\csname the#1\endcsname.\quad}
\renewcommand{\section}{\@startsection{section}{1}{\z@}{-2.2ex \@plus -.6ex \@minus -.2ex}{1.0ex \@plus .2ex}{\normalfont\fontsize{12}{14}\selectfont\bfseries}}
\renewcommand{\subsection}{\@startsection{subsection}{2}{\z@}{-1.8ex \@plus -.5ex \@minus -.2ex}{.8ex \@plus .2ex}{\normalfont\fontsize{11}{13}\selectfont\bfseries}}
\renewcommand{\subsubsection}{\@startsection{subsubsection}{3}{\z@}{-1.4ex \@plus -.4ex \@minus -.2ex}{.6ex \@plus .2ex}{\normalfont\normalsize\bfseries}}
\long\def\@makecaption#1#2{%
  \vskip\abovecaptionskip
  {\normalfont\normalsize\noindent #1\enspace|\enspace #2\par}%
  \vskip\belowcaptionskip
}
\makeatother
\newsavebox{\paperabstractcontent}
\newenvironment{paperabstract}{%
  \begin{lrbox}{\paperabstractcontent}%
  \begin{minipage}{\dimexpr\linewidth-20pt\relax}%
  \fontsize{10}{14.4}\selectfont\setlength{\parindent}{0pt}\setlength{\parskip}{4pt}%
}{%
  \end{minipage}\end{lrbox}%
  \par\noindent\begin{tikzpicture}
  \node[fill=black!5,draw=none,rounded corners=7pt,inner sep=10pt,outer sep=0pt] {\usebox{\paperabstractcontent}};
  \end{tikzpicture}\par\vspace{9pt}%
}
\hypersetup{pdftitle={SkillDuet: Co-Evolving Models and Skills through Response Feedback},pdfauthor={Xingguang Ji; Ziqiang Dong; Yubo Zhang; Xi Leng; Ninghua Yang; Yang Yang; Guanjun Jiang},pdfsubject={Qwen Business Unit of Alibaba}}
\title{SkillDuet: Co-Evolving Models and Skills\\through Response Feedback}
\author{Xingguang Ji, Ziqiang Dong, Yubo Zhang, Xi Leng, Ninghua Yang, Yang Yang, and Guanjun Jiang\\[3pt]
\normalfont\fontsize{8}{10}\selectfont Qwen Business Unit of Alibaba}\date{}
\begin{document}
\maketitle
\begin{paperabstract}
\textbf{Abstract.} During reinforcement learning, experience can be both internalized in LLM parameters and retained externally as reusable textual skills. Parameter updates alter the model's behavior and the interaction experience from which new skills are extracted. These skills, in turn, guide the model's subsequent interactions and parameter updates. As the model's capabilities evolve, previously beneficial skills may become redundant or detrimental. We present SkillDuet, a framework that coordinates parameter updates and skill evolution through model response feedback. Skills extracted from on-policy trajectories guide parameter updates through OPD at every step. A response-based score, Strength, ranks candidate skills by the magnitude of the shift they induce in the current model's responses, and the highest-scoring skills are periodically consolidated into a persistent global skill set (Global). The updated model and Global jointly guide subsequent interactions, generating experience for further parameter and skill updates. With Qwen3-4B, SkillDuet achieves success rates of 98.57\% on ALFWorld and 92.97\% on WebShop, exceeding the matched SEED-API baseline by 2.86 and 5.47 percentage points, respectively. Ablations show that response-based skill selection and Global consolidation improve task performance. Further experiments show that the benefits of retained skills depend on both their content and their compatibility with the model. SkillDuet provides a concrete mechanism for model--skill co-evolution and skill-level recursive self-improvement.\par\smallskip\textbf{Keywords:} self-evolving agents; skill evolution; on-policy distillation; skill internalization; skill consolidation; recursive self-improvement.
\end{paperabstract}

\section{Introduction}

\begin{figure}[tbp]\centering\includegraphics[width=\textwidth]{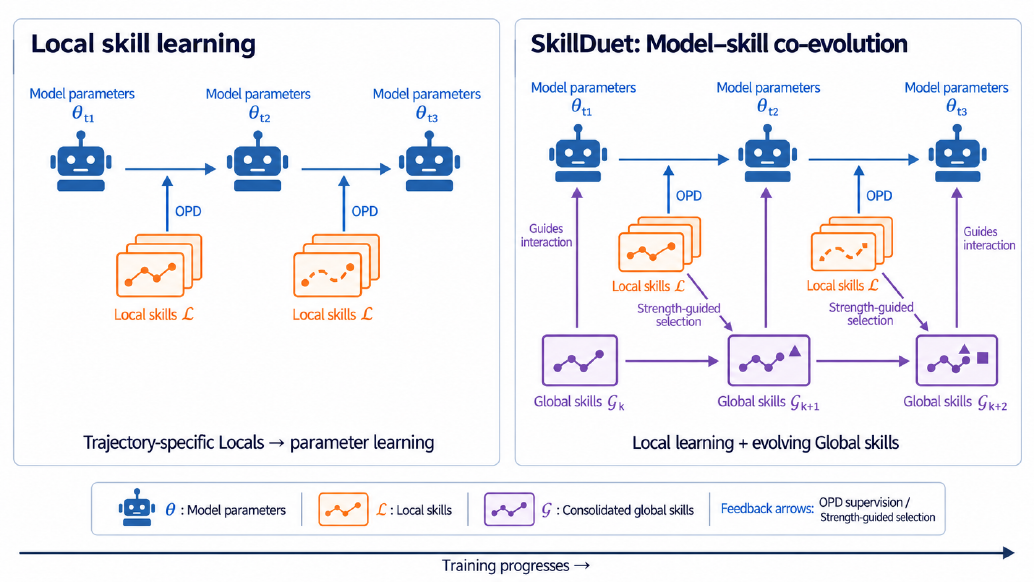}\caption{From Local-guided policy learning to model--skill co-evolution. Left: trajectory-specific Local skills supervise parameter updates through OPD. Right: SkillDuet also selects recent Local skills with Strength and consolidates them into Global, which guides subsequent interactions. Blue denotes model learning, amber denotes Local skills, and purple denotes Global skills.}\label{fig:teaser}\end{figure}

LLM agents can retain interaction experience through two routes: training-free approaches accumulate experience as external textual skills that guide later behavior, whereas training-based approaches internalize it into model parameters through SFT or RL. Textual skills can play both roles at once, serving as supervision for parameter updates and as guidance for future tasks. Recent work extracts retrievable, editable skills from interaction trajectories or distills their guidance into model parameters during RL \cite{skillrl,skillglow,seed}. Other methods consolidate lessons across trajectories or refine skill documents using execution feedback \cite{trace2skill,skillopt}.

However, a skill that provides useful supervision for the current parameter update may not remain useful as guidance for the updated model's subsequent interactions. A skill may become redundant once the model acquires the capabilities it supports. Other skills may remain useful, but their relevance can decline as the model's behavior changes. Retained skills may also conflict with newly learned behavior or existing guidance. Retained skills therefore need to be reassessed throughout training in light of both their content and the model's changing capabilities. Coordinating skill selection and refinement with parameter updates is central to skill-level recursive self-improvement (RSI), yet the two routes are usually studied separately. Coupling them is difficult: the model changes at every step, judging a skill's value through task execution is expensive, and the training objective itself provides no criterion for deciding which skills remain worth retaining.

Two observations from our experiments motivate this need for reassessment. On ALFWorld, guidance extracted from earlier trajectories substantially improves task completion early in training. When the guidance remains fixed, its gains diminish as training progresses, with a negative mean gain in the final stage (Section 3.1). In a separate evaluation, the same learned global skill set has different effects when supplied to two trained models (Section~\ref{sec:global_compatibility}). These observations show that the value of retained skills depends on both their content and their compatibility with the current model.

We introduce \textbf{SkillDuet}, a framework that coordinates parameter updates and skill evolution through \textbf{model response feedback} (Figure~\ref{fig:teaser}). SkillDuet couples continuous internalization with periodic consolidation. The agent acts with guidance from its global skill set (\textbf{Global}), and an external analyzer extracts local skills (\textbf{Local}) from the resulting trajectories. Local skills continuously provide privileged context for on-policy distillation (OPD), yielding token-level supervision for parameter updates at every step. We extend response feedback to long-term skill retention through \textbf{Strength}, a score that ranks recent Local skills by the magnitude of the shift they induce in the current model's responses. The selected skills are periodically consolidated with the existing Global. The updated model and Global jointly guide subsequent interactions, shaping the experience available for further parameter updates and skill extraction, so that internalization and consolidation reinforce each other. This feedback loop implements skill-level RSI through model--skill co-evolution.

We evaluate SkillDuet with Qwen3-4B on ALFWorld and WebShop, where it achieves success rates of 98.57\% and 92.97\%, respectively. These results exceed the matched SEED-API baseline by 2.86 and 5.47 percentage points. Ablation studies demonstrate the benefits of response-based Local selection and Global consolidation. Further comparisons identify the best-performing Local retention ratio among those tested and show how the content of retained skills and their compatibility with the model affect task performance.

\begin{samepage}Our main contributions are twofold:

\begin{itemize}

\item \textbf{A framework for model--skill co-evolution.} SkillDuet coordinates parameter updates and persistent skill updates within a single interaction loop. The evolving model and its skills jointly shape the experience used for subsequent learning.

\item \textbf{Response-guided skill evolution.} We introduce an approach to skill evolution that extends model response feedback from immediate distillation to long-term skill retention. Strength selects Local skills for Global consolidation based on the shift they induce in the current model's responses.

\end{itemize}

\end{samepage}

\begin{figure}[tb]\centering\includegraphics[width=.62\textwidth]{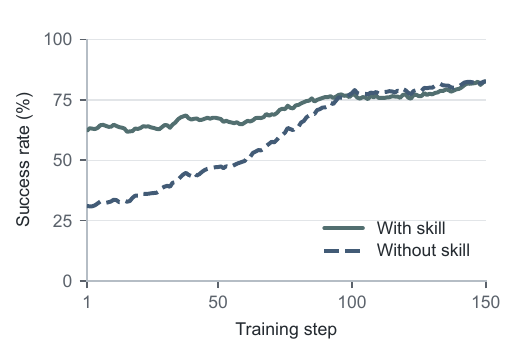}\caption{Effect of fixed skill guidance during GRPO training on ALFWorld. The curves show success rates of unguided training rollouts and additional skill-guided rollouts from the same model at each step. The guidance remains fixed while its benefit decreases.}\label{fig:guidance}\end{figure}

\section{Related Work}

\textbf{Skill construction and evolution.} Execution traces reveal reusable procedures as well as recurring mistakes. SkillRL maintains hierarchical skill libraries alongside reinforcement learning and recursive skill updates \cite{skillrl}. Trace2Skill analyzes a frozen agent's successful and failed trajectories in parallel, then hierarchically consolidates the resulting edits into transferable skills \cite{trace2skill}. SkillOpt proposes bounded edits through minibatch reflection, accepts them based on validation performance, and reuses rejected edits as feedback \cite{skillopt}. D2Skill combines success--failure comparisons with dynamic skill banks \cite{d2skill}. These methods use interaction outcomes to extract or refine external procedural guidance.

\textbf{Persistent knowledge and skill consolidation.} Retaining experience across updates requires deciding what to store and how to combine it. WikiSkill separates raw trajectories, persistent wiki knowledge, and executable skills. Behavioral patterns from past experience guide new proposals, while knowledge and edit history are preserved even when a skill update is rolled back \cite{wikiskill}. SkillGLoW groups experiences that share a solution procedure, compresses them into general guidance, and evaluates the updated guidance through execution; the agent then combines it with task-specific information \cite{skillglow}. Procedural Memory Distillation combines online reflection with batch experience and slower procedural consolidation \cite{pmd}. We build on this view of skills as persistent procedural knowledge and study how the current policy's responses can guide the choice of new experience to consolidate.

\textbf{Skill internalization and policy distillation.} SEED extracts hindsight skills with the policy model during RL and uses skill-conditioned response scores for on-policy distillation \cite{seed}. OPD-Evolver organizes distillation around a multilevel experience memory, while UCOB studies bidirectional distillation with and without skill conditioning, together with skill-utility updates \cite{opdevolver,ucob}. ATOD introduces annealed turn-aware distillation for multi-turn agents \cite{atod}. These methods convert textual or teacher-provided guidance into policy updates. SkillDuet retains this use of skills for immediate training and also uses model response feedback to select skills for persistent reuse.

\textbf{Skill management during policy adaptation.} Skill0.5 distinguishes skills used for internalization from those used during execution; SLIM uses a skill's contribution to execution to decide whether to retain, retire, or expand it \cite{skill05,slim}. UniSkill also evaluates skill proposals using the current actor's responses. It compares candidate and retrieved skills through action log-likelihood gaps on held-out successful and failed trajectories from the same task, and uses the signed success--failure difference to train the proposer and guide skill edits \cite{uniskill}. SkillDuet instead ranks recent Local skills by the magnitude of the response shift they induce and synthesizes the selected skills into Global. All eligible Local skills remain available for OPD, independently of this selection.

SkillDuet focuses on coordinating policy optimization and persistent skill updates as training proceeds. The current policy provides feedback for skill selection, and the retained skills guide future interactions. This two-way dependence allows skills to adapt as the model's capabilities change.

\section{Motivation and Method}

\subsection{Motivation: skill utility changes during training}

\textbf{Measuring the effect of fixed skills.} We train the model on ALFWorld for 150 steps using Group Relative Policy Optimization (GRPO). Before training, we extract a set of skills from the base model's trajectories and keep it fixed throughout the run (Appendix A). At each step, the same pre-update model generates eight unguided rollouts per task for the GRPO update and four additional rollouts with the fixed skills added to the input. These four skill-guided rollouts are used only for comparison and do not contribute to parameter updates. The success-rate difference between the guided and unguided rollouts therefore measures the execution-time effect of the fixed skills at that policy state.

\textbf{The same guidance becomes less useful as training proceeds.} Mean gains are +26.98, +9.64, and -2.00 percentage points over steps 1--50, 51--100, and 101--150, respectively (Figure~\ref{fig:guidance}). Thus, guidance that substantially helps early in training can become unhelpful later, even though the skill text has not changed. This result motivates updating the retained skills as the policy learns.

\begin{figure}[tbp]\centering\includegraphics[width=\textwidth]{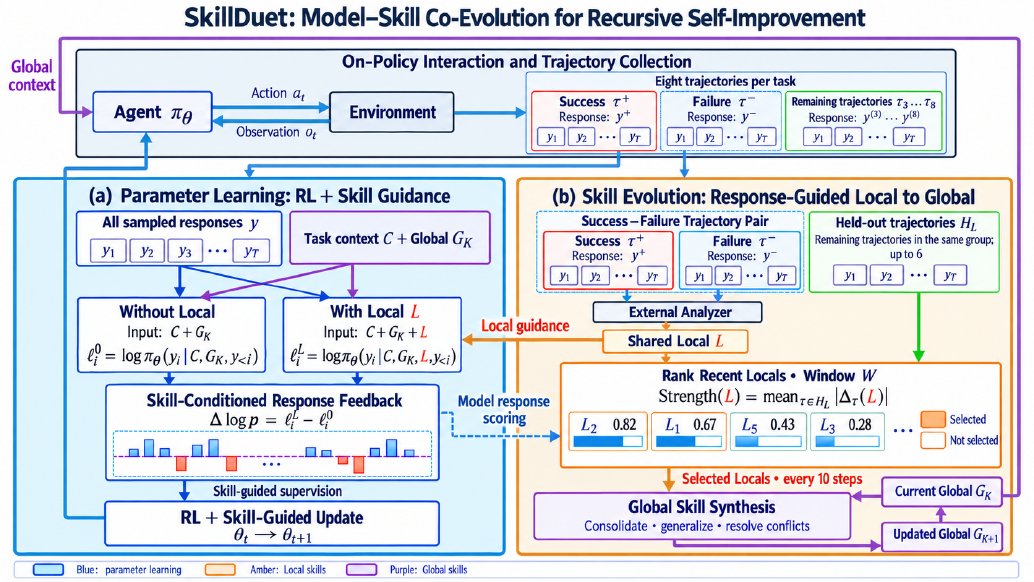}\caption{SkillDuet couples parameter learning with skill evolution. The agent collects eight trajectories per task under the current policy and Global. An external analyzer compares an eligible success--failure pair to produce a shared Local skill. (a) Local-conditioned response scores provide token-level OPD supervision alongside the RL objective. (b) Strength scores the Local skill on held-out trajectories; selected skills from a ten-step window are synthesized with the current Global. The updated policy and Global guide subsequent interactions. The dashed link denotes model response scoring for selection; the solid supervision path updates model parameters. Candidate scores illustrate ranking.}\label{fig:framework}\end{figure}

\subsection{Framework overview}

Figure~\ref{fig:framework} presents two coupled uses of interaction experience: parameter learning and skill evolution. The agent acts with its global skills (Global) in the prompt and collects eight trajectories per task. An external analyzer extracts local skills (Local) from these trajectories, capturing procedures or corrections grounded in the agent's recent behavior. Throughout, each Local refers to a single recently extracted skill, while Global refers to the consolidated set that persists across updates.

In the parameter-learning path (Figure~\ref{fig:framework}a), the agent scores sampled responses with and without Local guidance. These conditional scores provide token-level OPD supervision alongside the RL objective. All eligible Local skills can support the current policy update, regardless of their rank or whether they are later selected for Global consolidation (Sections 3.3--3.4).

In the skill-evolution path (Figure~\ref{fig:framework}b), an eligible success--failure pair yields a shared Local skill; groups without such a pair instead yield one Local per trajectory (Section 3.3). The remaining eligible trajectories provide held-out responses for Strength scoring (Section 3.5). The highest-scoring of the recent skills are periodically synthesized with the current Global to produce guidance reusable across task instances (Section 3.6). The updated policy and Global then guide further interactions, producing new experience for both learning paths.

Let $\theta$ denote the model parameters, $G$ the Global, and $\mathcal W$ the recent Local candidate window; initially, $G_0=\varnothing$. For task context $C$, $\pi_\theta$ denotes the model's conditional response distribution. We maintain one Global per task category on ALFWorld and a shared Global for all tasks on WebShop. The extraction, scoring, and synthesis procedures stay fixed, while the policy, skill content, and resulting response feedback change during training.

\subsection{Extracting Local skills from on-policy trajectories}

The trajectory group collected at each training step supplies the evidence for Local extraction (Figure~\ref{fig:framework}b). Environment feedback determines which of the eight trajectories succeed. If the group contains eligible successful and failed trajectories, an external Qwen3.8-27B analyzer compares one of each and generates a shared Local skill. Otherwise, it analyzes each eligible trajectory individually and generates one Local skill per trajectory. Appendix A specifies the trajectory and skill eligibility checks.

\textbf{Learning from success--failure pairs.} We select the first eligible successful trajectory and the first eligible failed trajectory in trajectory order. The analyzer receives only the task description and the observation--response histories of these two trajectories, identifies the decisions that distinguish them, and produces one shared Local skill. This comparison grounds corrective guidance in successful behavior that the current policy has already demonstrated. The shared skill provides OPD supervision for all eligible trajectories in the group, including its two source trajectories. We score its usefulness for Global consolidation on the remaining, held-out trajectories, of which there are at most six.

\textbf{Learning from individual trajectories.} If no eligible success--failure pair exists, the analyzer processes each eligible trajectory using the task description and its observation--response history. It extracts reusable procedures from successes and guidance for avoiding or correcting failures. Each resulting Local skill supervises OPD on its own source trajectory and does not enter Global consolidation.

\subsection{Updating model parameters with Local-guided OPD}

In the parameter-learning path (Figure~\ref{fig:framework}a), the external analyzer provides Local text, while the agent model supplies the teacher log probabilities used for distillation. For a sampled response, the student branch computes token log probabilities under context $C$ and Global $G$. Adding a Local skill $L$ to this context produces the teacher branch, which scores the same response tokens. Their log-probability difference provides a token-level OPD signal, and a confidence gate weights each token's supervision according to this difference \cite{seed}. Let $s_i$ and $t_i$ denote the student and teacher token log probabilities at token position $i$, $M_i$ the supervision mask, and $\operatorname{sg}$ the stop-gradient operator:

\[
\begin{aligned}
g_i&=\operatorname{sg}[\sigma(\beta(t_i-s_i))],\quad \beta=5,\\
\mathcal L_{\mathrm{OPD}}&=\frac{\sum_i M_i g_i(\operatorname{sg}[t_i]-s_i)}{\sum_i M_i},\\
\mathcal L&=\mathcal L_{\mathrm{PPO}}+\lambda_{\mathrm{OPD}}\mathcal L_{\mathrm{OPD}}+\mathcal L_{\mathrm{reg}}.
\end{aligned}
\tag{1}
\]

The policy is optimized with the combined objective in Equation (1). Proximal policy optimization (PPO) supplies the outcome-based RL objective \cite{ppo}, while $\mathcal L_{\mathrm{reg}}$ contains the active regularization terms. Using an external Qwen3.8-27B analyzer allows skill extraction without requiring the policy itself to analyze trajectories. Appendix A describes how teacher scores are cached and aggregated.

\subsection{Selecting Local skills with model response feedback}

A persistent skill set should retain experience that remains useful to the current policy. If a Local skill barely changes the model's response probabilities, its guidance may already be reflected in the model's parameters. A larger change may instead identify information worth retaining. This motivates Strength, a score used to rank recent Local skills by the magnitude of the response shift they induce.

The skill-evolution path in Figure~\ref{fig:framework}b evaluates each Local on a scoring set $H_L$ of held-out trajectories that were not used to generate it. Scoring keeps the model parameters, interaction context, sampled responses, and Global fixed. Let $\Delta_r(L)$ denote the mean response log-probability shift over the valid tokens of trajectory $r$. Appendix A specifies the scoring reference and its implementation.

Strength takes the absolute value of each trajectory-level mean and then averages these values, giving equal weight to each scoring trajectory:

\[
\operatorname{Strength}(L)=\frac{1}{|H_L|}\sum_{r\in H_L}|\Delta_r(L)|.
\tag{2}
\]

Each scorable Local skill is evaluated before the parameter update at its generation step. Its Strength score is cached at generation and used to rank candidates at the end of the window; each score therefore reflects the policy at the corresponding skill's generation step. Scoring reuses sampled trajectories and requires no additional environment interaction. Appendix A, Algorithm 1, gives the full procedure.

\subsection{Periodically consolidating Local skills into Global}

Skill selection and Global consolidation operate every ten training steps (Figure~\ref{fig:framework}b). SkillDuet deduplicates the recent Local candidates for each Global and retains the top 50\% by cached Strength. The synthesizer receives these skills together with the current Global. If the input exceeds the context budget, hierarchical compression first combines shared procedures and preserves the conditions under which they apply. The synthesizer then integrates the selected experience with existing rules to produce compact, reusable guidance (Figure~\ref{fig:case}; Appendix G).

After the update, we clear the temporary Local window and use the new Global in subsequent agent prompts. The next training window therefore collects experience under both an updated policy and updated skill guidance.

\begin{samepage}

\section{Experiments}

We evaluate whether coordinating policy optimization with skill consolidation improves agent performance on ALFWorld and WebShop. Training ablations examine Global consolidation, response-based Local selection, the selection criterion, and the retention ratio. We then hold model parameters fixed to test how Global's content and its compatibility with the model affect task execution.

\end{samepage}

\begin{table}[tbp]\centering
\caption{Performance of Qwen3-4B-family agents on ALFWorld and WebShop (\%).}\label{tab:1}
\fontsize{8.5pt}{10.3pt}\selectfont\setlength{\tabcolsep}{2pt}\renewcommand{\arraystretch}{1.16}
\begin{tabular}{@{}>{\raggedright\arraybackslash}p{0.27\dimexpr\textwidth-36pt\relax}>{\raggedleft\arraybackslash}p{0.075\dimexpr\textwidth-36pt\relax}>{\raggedleft\arraybackslash}p{0.075\dimexpr\textwidth-36pt\relax}>{\raggedleft\arraybackslash}p{0.075\dimexpr\textwidth-36pt\relax}>{\raggedleft\arraybackslash}p{0.075\dimexpr\textwidth-36pt\relax}>{\raggedleft\arraybackslash}p{0.075\dimexpr\textwidth-36pt\relax}>{\raggedleft\arraybackslash}p{0.075\dimexpr\textwidth-36pt\relax}>{\raggedleft\arraybackslash}p{0.09\dimexpr\textwidth-36pt\relax}>{\raggedleft\arraybackslash}p{0.095\dimexpr\textwidth-36pt\relax}>{\raggedleft\arraybackslash}p{0.095\dimexpr\textwidth-36pt\relax}@{}}
\toprule
 & \multicolumn{7}{c}{\textbf{ALFWorld}} & \multicolumn{2}{c}{\textbf{WebShop}} \\
\cmidrule(lr){2-8}\cmidrule(lr){9-10}
\textbf{Method} & \textbf{Pick $\uparrow$} & \textbf{Look $\uparrow$} & \textbf{Clean $\uparrow$} & \textbf{Heat $\uparrow$} & \textbf{Cool $\uparrow$} & \textbf{Pick2 $\uparrow$} & \textbf{ALL $\uparrow$} & \textbf{Score $\uparrow$} & \textbf{Succ. $\uparrow$} \\
\midrule
SkillRL (O3) \cite{d2skill} & \underline{90.0} & \underline{63.6} & \underline{92.3} & \underline{42.9} & \underline{52.0} & \underline{40.9} & \underline{67.2} & \underline{89.4} & \underline{75.7} \\
D2Skill (O3) \cite{d2skill} & \underline{89.4} & \underline{54.5} & \underline{72.4} & \underline{60.0} & \underline{66.7} & \underline{50.0} & \underline{72.7} & \underline{89.1} & \underline{75.8} \\
D2Skill (Gemini-3-Flash) \cite{d2skill} & \underline{88.6} & \underline{66.7} & \underline{75.0} & \underline{60.0} & \underline{54.2} & \underline{52.6} & \underline{69.6} & \underline{90.0} & \underline{82.8} \\
D2Skill + teacher SFT \cite{d2skill} & \underline{97.6} & \underline{88.9} & \underline{95.8} & \underline{90.0} & \underline{100.0} & \underline{91.7} & \underline{95.3} & \underline{89.2} & \underline{81.3} \\
OPD (ATOD) \cite{atod} & \textemdash & \textemdash & \textemdash & \textemdash & \textemdash & \textemdash & \underline{80.47} & \textemdash & \underline{75.00} \\
GSPO (ICRL) \cite{icrl} & \textemdash & \textemdash & \textemdash & \textemdash & \textemdash & \textemdash & \underline{82.0} & \underline{82.1} & \underline{65.0} \\
Vanilla & 70.83 & 40.00 & 19.75 & 18.52 & 20.00 & 26.67 & 37.38 & 13.85 & 0.00 \\
GRPO & 100.00 & 85.71 & 94.29 & 61.54 & 82.76 & 72.22 & 86.43 & 92.96 & 89.84 \\
SDAR & 100.00 & 100.00 & 100.00 & 54.55 & 70.37 & 80.95 & 87.86 & 89.83 & 86.72 \\
SEED & 100.00 & 100.00 & 100.00 & 100.00 & 90.48 & 87.50 & 96.09 & 86.23 & 75.78 \\
SEED-API & 100.00 & 93.33 & 96.30 & 83.33 & 95.00 & 100.00 & 95.71 & 91.31 & 87.50 \\
\textbf{SkillDuet} & \textbf{100.00} & \textbf{100.00} & \textbf{100.00} & \textbf{100.00} & 90.00 & \textbf{100.00} & \textbf{98.57} & \textbf{95.18} & \textbf{92.97} \\
\bottomrule\end{tabular}
\par\smallskip
\noindent Underlined entries are taken from the cited papers and follow their original evaluation and aggregation protocols (Appendix B). ALL is overall ALFWorld success; Score is WebShop's terminal reward. Dashes indicate unreported results. OPD (ATOD) and GSPO (ICRL) denote the OPD and GSPO results reported in the ATOD and ICRL studies, respectively. Appendix E details checkpoint selection, category averages, and evaluation denominators. Vanilla is Qwen3-4B without task-specific training. Its evaluation sizes are given in Appendix E.
\end{table}

\subsection{Experimental setup}

\textbf{Benchmarks and metrics.} We use Qwen3-4B on ALFWorld household tasks \cite{alfworld} and WebShop shopping tasks \cite{webshop}. For ALFWorld, Table 1 reports success rates for six task categories and overall success (ALL). For WebShop, we report mean continuous terminal reward (Score) and complete-task success rate (Succ.), using catalog1000 and its corresponding document index.

\textbf{Baselines.} Our matched baseline, SEED-API, retains SEED's per-trajectory skill extraction and OPD pipeline, uses the same external analyzer as SkillDuet, and does not maintain Global. Both methods start from the same Qwen3-4B checkpoint without task-specific supervised fine-tuning (SFT). Other baselines include GRPO \cite{grpo}, SDAR \cite{sdar}, our Qwen3-4B reproduction of SEED, and Vanilla Qwen3-4B without task-specific training. We also include published results for Qwen3-4B-family models. Full configurations and evaluation protocols are provided in Appendices A, B, and E.

\textbf{Training.} SkillDuet starts from Qwen3-4B without SFT and runs for 300 steps. The OPD coefficient is .05 on ALFWorld and .03 on WebShop. Sampling, window management, and optimization settings are given in Appendix A.

\subsection{Main results}

\textbf{SkillDuet improves on the matched SEED-API baseline.} With the external analyzer and initial checkpoint held constant, SkillDuet increases ALFWorld success from 95.71\% to 98.57\% and WebShop success from 87.50\% to 92.97\%. These are gains of 2.86 and 5.47 percentage points, respectively.

\textbf{The gains extend to both task completion and goal matching.} SkillDuet reaches 100\% success in five ALFWorld categories. On Heat, success rises from SEED-API's 83.33\% to 100\%, showing improved completion of tasks that require multiple ordered actions. On WebShop, Score rises from 91.31 to 95.18 alongside the increase in complete-task success. Compared with GRPO, SkillDuet improves Score by 2.22 points and success by 3.13 percentage points.

\textbf{Performance before task-specific training.} Vanilla Qwen3-4B achieves 37.38\% success on ALFWorld and a WebShop Score of 13.85 with 0.00\% complete-task success. These values provide a reference for the capability of the initial model before adaptation.

\subsection{Global consolidation and response-based selection}\label{sec:global_selection}

We test whether Global consolidation improves on using Local skills only for parameter updates, and whether model response feedback improves the choice of skills to consolidate.

On WebShop, SEED-API extracts one Local skill per trajectory for OPD. Paired Local-only instead generates a shared skill from a successful and a failed trajectory of the same task, while still training without Global. Comparing these two configurations isolates the Local-generation strategy. Full SkillDuet adds response-based selection and Global consolidation to Paired Local-only, testing their combined contribution.

To isolate the role of response feedback, we compare Strength with Random while keeping the candidate window, retention ratio, synthesis procedure, and Global deployment fixed. Both collect eligible Local skills over ten steps for each Global, remove duplicates, and retain half of the candidates. Strength ranks candidates by the magnitude of their effect on model responses; Random selects them without response feedback.

\par\smallskip\noindent\begin{minipage}{\columnwidth}\small\setlength{\tabcolsep}{3pt}\renewcommand{\arraystretch}{1.10}
\inlinecaption{table}{WebShop ablations of Global consolidation and Local selection. Values are peak validation success rates (\%).}{tab:2}
\begin{tabular}{@{}p{0.6\dimexpr\columnwidth-12pt\relax}p{0.18\dimexpr\columnwidth-12pt\relax}p{0.22\dimexpr\columnwidth-12pt\relax}@{}}
\toprule
\textbf{Configuration} & \textbf{Global} & \textbf{Succ. $\uparrow$} \\
\midrule
SEED-API & None & 87.50 \\
Paired Local-only & None & 87.50 \\
SkillDuet: Random & Top50\% & 89.06 \\
\textbf{SkillDuet: Strength} & Top50\% & \textbf{92.97} \\
\bottomrule\end{tabular}
\end{minipage}\par\smallskip

SEED-API and Paired Local-only both reach 87.50\% peak validation success, so paired generation alone does not improve this metric. Full SkillDuet reaches 92.97\%, a gain of 5.47 percentage points over either configuration. Random selection reaches 89.06\%, trailing Strength by 3.91 percentage points (Table 2).

Consolidating selected Local skills into Global improves performance beyond Local-guided parameter updates alone. Response-based selection also outperforms random selection.

\subsection{Response-based criteria for Local selection}\label{sec:selection_criteria}

We next examine which response-based criterion best supports Local skill selection: response-shift magnitude, reward weighting, or success--failure contrast.

We compare Strength, Utility, Contrast, and Random in a historical ALFWorld candidate pool. Each configuration is trained for 300 steps using the skills selected by its criterion. We measure peak validation success and normalized area under the validation curve (AUC). Appendix C provides details of pool management and score accumulation.

Strength measures the magnitude of a Local-induced response shift. Utility weights the shift by trajectory advantage, and Contrast compares shifts on successful and failed trajectories. Random assigns priorities independently of model responses. The three response-based scores are:

\[
\begin{aligned}
S(L)&=\frac{1}{|H_L|}\sum_{r\in H_L}|\Delta_r(L)|,\\
U(L)&=\frac{1}{|H_L|}\sum_{r\in H_L}A_r\Delta_r(L),\\
C(L)&=\frac{1}{|H_L^+|}\sum_{r\in H_L^+}\Delta_r(L)\\
&\quad-\frac{1}{|H_L^-|}\sum_{r\in H_L^-}\Delta_r(L).
\end{aligned}
\tag{3}
\]

Here, $\Delta_r(L)$ is the trajectory-mean response shift defined in Section 3.5; $H_L$ is the scoring set; $A_r$ is the trajectory-mean training advantage; and $H_L^+$ and $H_L^-$ are the successful and failed subsets of $H_L$. Contrast falls back to Utility when the scoring set does not support a success--failure comparison.

Strength achieves the highest peak validation success rate, 97.9\%, and the highest AUC, 88.26\%, among the four criteria. Its AUC exceeds Random by 9.29 percentage points (Figure~\ref{fig:signals}).

\par\medskip\noindent\begin{minipage}{\textwidth}\centering\includegraphics[width=\textwidth]{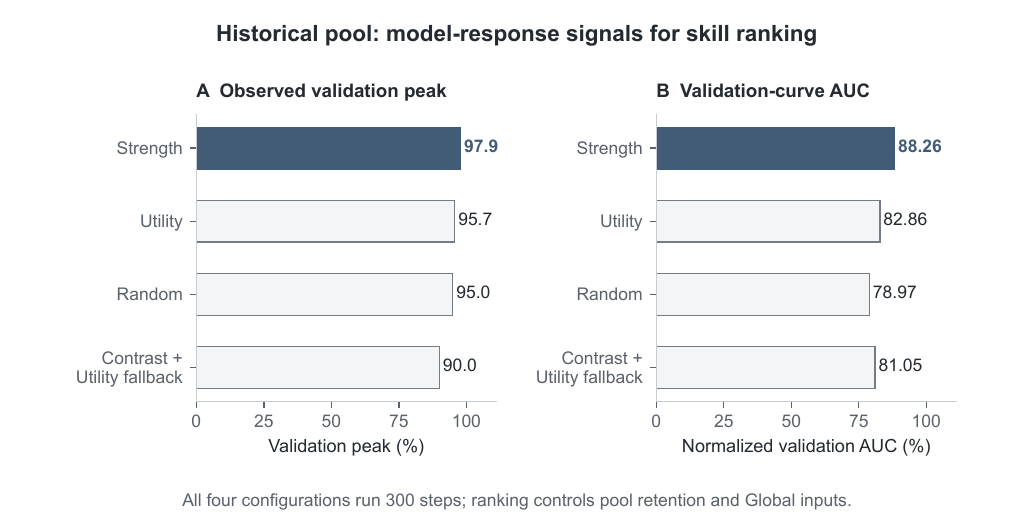}\inlinecaption{figure}{Local-selection criteria in the historical ALFWorld candidate pool. Strength achieves the highest peak validation success (left) and normalized area under the validation curve (AUC; right).}{fig:signals}\end{minipage}\par\medskip

Trajectory rewards and success labels describe behavior before the candidate skill is introduced; they do not directly measure the skill's value for future execution. Increasing the likelihood of a successful response, or decreasing that of a failed one, likewise does not establish that the skill will improve later decisions. Weighting response shifts by the original trajectory's quality can therefore mix trajectory quality with the effect of the skill.

In this study, response-shift magnitude is more effective for skill selection than reward weighting or success--failure contrast. These results support Strength as a practical criterion for long-term skill retention, without treating the direction of the probability shift as a direct measure of benefit.

\subsection{Local retention ratio}\label{sec:retention_ratio}

We test how the fraction of Local skills retained for Global consolidation affects task performance. Retaining too few candidates may discard useful procedures, whereas retaining too many may introduce redundancy or conflicting guidance.

We compare Top25\%, Top50\%, and Top75\% retention on ALFWorld and WebShop, selecting candidates in descending Strength order for each Global. All eligible Local skills continue to support OPD, so this comparison changes only the experience used for Global consolidation. We report peak validation success during training; evaluation details and exact values are given in Appendices A and D.

Top50\% matches Top25\% on ALFWorld and achieves the highest result on WebShop, while Top75\% performs worse than Top50\% on both tasks (Figure~\ref{fig:selection}). Top50\% therefore provides the best overall performance among the tested ratios and is our default.

\par\medskip\noindent\begin{minipage}{\textwidth}\centering\includegraphics[width=\textwidth]{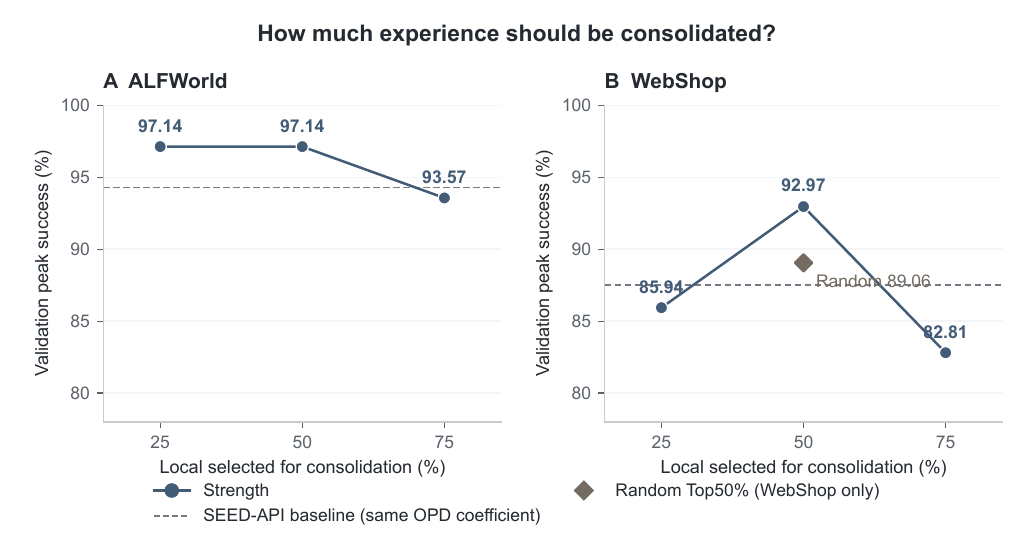}\inlinecaption{figure}{Effect of the Local retention ratio on peak validation success. Top50\% matches Top25\% on ALFWorld and achieves the best WebShop result. The Random baseline uses the same Top50\% consolidation and Global deployment procedure.}{fig:selection}\end{minipage}\par\medskip

\subsection{Global content and model compatibility}\label{sec:global_compatibility}

We examine whether the trained model retains task capability without Global, whether Global's content contributes to performance, and whether the same skills benefit a different trained model. All comparisons hold model parameters fixed.

On ALFWorld, removing Global from SkillDuet tests the model's capability without external skill guidance. Replacing Global with length-matched neutral text tests the role of skill content while controlling for prompt position, format, and length. Supplying SkillDuet's Global to SEED-API tests whether its benefit transfers to another trained model. All conditions use matched evaluation tasks, and both conditions with learned Global use the same SkillDuet skill set.

\par\smallskip\noindent\begin{minipage}{\columnwidth}\small\setlength{\tabcolsep}{3pt}\renewcommand{\arraystretch}{1.10}
\inlinecaption{table}{Controlled ALFWorld evaluations with fixed model parameters (\%). ALL is overall success; Avg is the unweighted mean across the six task categories.}{tab:3}
\begin{tabular}{@{}p{0.29\dimexpr\columnwidth-18pt\relax}>{\raggedleft\arraybackslash}p{0.29\dimexpr\columnwidth-18pt\relax}>{\raggedleft\arraybackslash}p{0.21\dimexpr\columnwidth-18pt\relax}>{\raggedleft\arraybackslash}p{0.21\dimexpr\columnwidth-18pt\relax}@{}}
\toprule
\textbf{Model} & \textbf{Global} & \textbf{ALL $\uparrow$} & \textbf{Avg $\uparrow$} \\
\midrule
Qwen3-4B (Vanilla) & None & 37.38 & 32.63 \\
SEED-API & None & 93.81 & 92.43 \\
SEED-API & Learned Global & 89.76 & 88.42 \\
SkillDuet & None & 76.67 & 73.72 \\
SkillDuet & Length-matched neutral & 75.24 & 72.17 \\
\textbf{SkillDuet} & \textbf{Learned Global} & \textbf{97.62} & \textbf{97.50} \\
\bottomrule\end{tabular}
\end{minipage}\par\smallskip

SkillDuet retains substantial task capability without Global, and its learned Global skills further improve performance (Table 3). Replacing Global with length-matched neutral text reduces performance, indicating that this gain depends on skill content. The same skills lower SEED-API's success rate, showing that their benefit does not transfer automatically to another trained model.

SkillDuet without Global, or with neutral text in its place, also performs below SEED-API without Global. This difference may partly reflect a change in input context: SkillDuet's parameter updates adapt the model to Global-conditioned prompts, whereas removal or replacement changes that context at evaluation.

These results show that Global's benefit depends on both skill content and its compatibility with the model, supporting the joint evolution of model parameters and skills. The paired Clean example below illustrates how a Global rule is reflected in the model's actions.

\par\medskip\noindent\begingroup\setlength{\fboxsep}{7pt}\setlength{\fboxrule}{.5pt}%
\fcolorbox{blue!40!black}{blue!3}{\begin{minipage}{\dimexpr\linewidth-2\fboxsep-2\fboxrule\relax}\small\raggedright\setlength{\parindent}{0pt}
\textbf{Paired execution example: clean before placing}\par\smallskip \textbf{Task.} Place a cleaned butter knife on the dining table.\par\smallskip \textbf{With Global.} Pick up the knife, go to the sink, clean it, and place it on the table; the task succeeds in 7 actions.\par\smallskip \textbf{Without Global.} Place the uncleaned knife first, then cycle through examination and pick-up/place actions; the task remains unfinished after 30 actions.\par\smallskip With the model parameters fixed, the successful trajectory follows Global's clean-before-place procedure. Appendix F provides the paired records and their provenance.
\end{minipage}}\endgroup\par\medskip

\begin{samepage}

\section{Conclusion}

We presented SkillDuet, a framework for coordinating LLM parameter updates and the evolution of reusable textual skills during RL. These processes are interdependent: skills guide model learning, while parameter updates change which skills remain useful. Model response feedback connects them in a shared learning loop. The evolving model and its skills jointly shape subsequent interactions and the experience available for further learning.

\end{samepage}

Experiments with Qwen3-4B on ALFWorld and WebShop show gains over the matched SEED-API baseline. Ablations support the benefits of response-based skill selection and consolidation. Further analyses show that a skill's benefit depends on its content and its compatibility with the model. SkillDuet provides a concrete mechanism for skill-level recursive self-improvement through model--skill co-evolution.

\clearpage\normalsize

\section*{Appendix A. Training and implementation details}

\subsubsection*{Training configurations}

We use Qwen3-4B as the backbone. SkillDuet starts RL without SFT, trains for 300 steps, and samples eight trajectories per task. The OPD coefficient is .05 for ALFWorld and .03 for WebShop. Each Global has a ten-step candidate window, from which the top 50\% of eligible Local skills are retained. SEED-API runs for 300 steps with the same task-specific OPD coefficients as SkillDuet. Both methods use confidence-gated OPD.

Our SEED reproduction uses SFT initialization followed by 150 RL steps with an OPD coefficient of .01. The external annotator (UDE in the implementation) generates the SFT skill targets; during RL, the policy analyzes its own trajectories without external analyzer calls.

GRPO provides the outcome-based RL baseline. Our reproduction of Self-Distilled Agentic Reinforcement Learning (SDAR) follows the official implementation and uses its static skill files. The student samples responses without skills, and the same model scores those responses under the original prompt augmented with static skills to provide detached teacher targets. Training combines GRPO with confidence-gated distillation, using a gate scale of five and a distillation coefficient of .01. SDAR runs for 300 steps without SFT, online skill generation, skill updates, or external analyzer calls.

\par\smallskip\noindent\begin{minipage}{\linewidth}\small\raggedright
\textbf{Algorithm 1. Local-guided policy updates and cached-score Global selection}\par\smallskip
\setlength{\tabcolsep}{0pt}\renewcommand{\arraystretch}{1.00}
\begin{tabular}{@{}r@{\hspace{7pt}}>{\raggedright\arraybackslash}p{\dimexpr\linewidth-22pt\relax}@{}}\toprule
1 & Input: initial parameters $\theta_0$; steps $T=300$, window $w=10$, ratio $\rho=.5$. \\
2 & Initialize $G\leftarrow\varnothing$ and temporary windows $\mathcal W_b\leftarrow\varnothing$ for all Globals. \\
3 & For training step $k=1,\ldots,T$: \\
4 & \hspace*{0.9em}If $k>1$ and $(k-1)\bmod w=0$: \\
5 & \hspace*{1.8em}For each Global indexed by $b$: \\
6 & \hspace*{2.7em}$E_b\leftarrow\mathcal W_b$ after deduplication and eligibility/valid-score checks. \\
7 & \hspace*{2.7em}If $E_b=\varnothing$, retain $G_b$ and continue to the next Global. \\
8 & \hspace*{2.7em}Read cached scores; retain the top $m_b=\max(1,\lceil\rho|E_b|\rceil)$ Locals in descending order. \\
9 & \hspace*{2.7em}Synthesize with $G_b$; activate the update if synthesis checks pass, otherwise retain $G_b$. \\
10 & \hspace*{1.8em}Clear temporary windows for all Globals and start the next window. \\
11 & \hspace*{0.9em}Set $\bar\pi_k=\pi_{\theta_{k-1}}$; sample eight trajectories per task with $G$. \\
12 & \hspace*{0.9em}Generate all eligible Locals as in Section 3.3 and retain them for current-step OPD. \\
13 & \hspace*{0.9em}For each success--failure Local $L$ with a nonempty independent scoring set $H_L$: \\
14 & \hspace*{1.8em}Score under $\bar\pi_k$ using Equation (2) and Appendix A; store text and score in $\mathcal W_{b(L)}$ and record provenance. \\
15 & \hspace*{0.9em}Optimize PPO and OPD with all eligible Locals in Equation (1), obtaining $\theta_k$. \\
\bottomrule\end{tabular}\end{minipage}\par\smallskip

\subsubsection*{Candidate windows and scoring inputs}

ALFWorld has one Global for each of its six task categories; WebShop uses one shared Global. Within each window, Local skills with identical text after whitespace normalization are deduplicated while their source records are retained. A skill is scored once when it is generated, before the policy update. At the window boundary, we read these cached scores, check candidate eligibility, and round the retention count upward, keeping at least one candidate from a nonempty eligible set. Consolidation occurs before the next rollout step. The temporary candidates and scores are then cleared.

For Strength scoring, the reference context uses token-length-matched neutral text $N_L$ in place of Local. With the policy and sampled response fixed, the trajectory-level shift is:

\[
\begin{aligned}
\Delta_r(L)&=\frac{1}{T_r}\sum_{i\in r}\Big[\\
&\quad\log\pi(a_i\mid C,G,L,a_{<i})\\
&\quad-\log\pi(a_i\mid C,G,N_L,a_{<i})\Big].
\end{aligned}
\tag{A1}
\]

To construct neutral text, we repeat or truncate a fixed filler passage containing no task instructions or procedures. We match the token counts of the complete, untruncated prompts while keeping the prompt template and interaction context fixed. The verified main WebShop configuration permits a difference of at most two tokens, and neither prompt may exceed its length limit. The scoring set $H_L$ excludes the skill's generation pair and retains only held-out trajectories with valid response tokens, scores under both conditions, and valid length matching. A Local skill without a valid score is not considered for Global selection.

\subsubsection*{Assigning Local skills to supervision trajectories}

Trajectory analysis groups records by trajectory identity and reconstructs the sequence of observations and responses. We decode responses using the generation mask and prefer raw environment observations when available. Terminal success indicators distinguish successful and failed trajectories; action validity is not used to filter the response text. For immediate OPD, a generated skill must contain a parsable, nonempty episode\_skill field with at least five whitespace-separated words, excluding punctuation-only strings and placeholders.

A per-trajectory Local skill supervises only its source trajectory. A skill generated from a success--failure pair supervises all eligible trajectories in the group. We choose the first eligible success and failure for generation and exclude both from Strength scoring. To enter Global selection, a shared Local skill must also receive a valid score on at least one held-out trajectory. Skills that cannot be scored can still provide immediate OPD supervision. Appendix C describes the separate historical-pool experiments.

The analyzer is instructed to produce actionable guidance grounded in the supplied trajectories. The paired prompt contains: ``Two trajectories of the SAME task are given below. Compare them, locate the earliest step where their decisions diverge, and write one episode\_skill that turns that divergence into a rule the policy can act on.'' The response schema is \{"episode\_skill": "string"\}. Single-trajectory prompts ask for reusable procedures from successes or guidance for avoiding and correcting failures.

\subsubsection*{Global synthesis and deployment}

Global starts empty and is first synthesized after the initial ten-step window. Local extraction and Global synthesis use the same external API client and model endpoint, with separate prompts for trajectory analysis and skill consolidation. Selected Local skills are merged with the current Global, either directly or after hierarchical compression (Figure~\ref{fig:case}). The generated skill set must be parsable, nonempty, and satisfy the required structure and length constraints. If these checks fail, we keep the existing Global. Appendix D describes the optional execution-based Commit check.

The recorded main WebShop synthesis configuration uses a context budget of 24,000, reserving 12,288 for output and 1,024 as a safety margin, leaving 10,688 for input. The implementation estimates this budget from UTF-8 byte counts rather than exact provider token counts. Larger inputs trigger hierarchical compression, with up to five levels. In the example from steps 11--20, the initial input estimate is 22,280 and the final synthesis input is 9,287.

Global contains at most six skills, each with at least five words, and is limited to 2,400 characters in total, with a target of about 45 words per skill. These values describe the recorded configuration.

The Global system prompt is: ``Synthesize evidence-linked procedural skills. Return JSON only.'' The user prompt asks the synthesizer to merge duplicate procedures, preserve conditions and exceptions, remove instance identifiers, and return the complete updated Global. Useful existing rules must be preserved verbatim even if the recent window contains no matching examples. New or revised rules must cite supplied source identifiers. The output schema is \{"skills": [\{"text": "conditional procedure", "source\_ids": ["L0000"]\}]\}. Intermediate hierarchical summaries also retain source identifiers and applicability conditions.

\subsubsection*{Teacher scores and implementation details}

The external analyzer supplies Local text, while the skill-conditioned agent supplies the teacher log probabilities. In the main WebShop configuration, teacher scores are cached before each actor update and refreshed at the next RL step. Student scores are recomputed during optimization. Both the teacher scores and the confidence gate are detached from the gradient. When the teacher--student difference is zero, the sigmoid gate equals 0.5; a zero loss value therefore does not imply a zero gradient.

The original SEED formulation evaluates both branches with the current model parameters during optimization. Complete deployment records are unavailable for some earlier and original configurations.

\subsubsection*{Optimization and runtime}

RL training uses a learning rate of 1e-6, weight decay of .01, no warmup, a constant learning-rate schedule, a training batch size of 16, and eight trajectories per task. Rollouts use temperature=1 and top-p=1. Validation samples one trajectory per task at temperature=.4, with a response limit of 512 tokens.

For ALFWorld, SkillDuet uses mini/micro batch sizes of 256/16, GRPO and SDAR use 64/8, and SEED uses 256/32. The main WebShop configurations use 64/8, and the main WebShop SkillDuet configuration uses AdamW. The configurations specify a single node with eight GPUs; the GPU model is not fully documented in the available records.

We refer to the external analyzer as Qwen3.8-27B. The verified synthesis requests use the endpoint alias edu-vlm-rm, which does not identify an immutable backend weight version for all earlier experiments. Runtime records for the main WebShop SkillDuet configuration list PyTorch 2.6.0, vLLM 0.8.5.post1, Transformers 4.51.3, and Ray 2.43.0.

\subsubsection*{Evaluation protocols}

The main ALFWorld configurations use 140 evaluation trajectories, while SEED with SFT uses 128. Validation is performed every thirty and five steps, respectively. The SkillDuet and SEED-API (.05) results in Table 1 come from separate evaluations of their step 300 checkpoints, each with category denominators of 40/15/27/18/20/20. Other results from our experiments use the checkpoint with the highest validation success within the training budget. The SEED row is our Qwen3-4B reproduction.

WebShop uses catalog1000 and its corresponding index, with training goal IDs at least 500 and validation IDs 0--499. We select checkpoints by validation Succ. and report Score from the same evaluation.

Each training configuration is represented by one training run. The three seeds in Tables 3 and 8 are repeated execution evaluations of fixed checkpoints, not independent training runs.

The ALFWorld retention-ratio study in Appendix D uses earlier Strength-window configurations with an OPD coefficient of .05 and 140-trajectory training-time validation. Top25\%, Top50\%, and Top75\% select steps 240, 300, and 300, respectively. The Top50\% value of 97.14\% comes from that validation sequence. The 98.57\% in Tables 1 and 6 comes from a later evaluation of the step 300 checkpoint with a new task order and environment RNG. Figure~\ref{fig:selection} also uses the earlier SEED-API (.05) validation reference, approximately 94.3\% at step 240; Table 1 instead reports the separate step 300 evaluation of 95.71\%.

\textbf{Fixed-skill probe.} For the 150-step GRPO training run in Section 3.1, we select episode\_skill records from the base model's trajectories before training, group them by category, and remove duplicates. The resulting fixed guidance covers the six ALFWorld categories and default rules and remains unchanged throughout training. At each step, the same pre-update model generates eight unguided rollouts per task for the GRPO update and four additional rollouts with the fixed skills added to the input. The four skill-guided rollouts are used only for comparison and do not enter optimization. We average task-paired success-rate differences over consecutive fifty-step intervals.

\subsubsection*{Computation and API costs}

The available records for the main WebShop run contain 35,036 Local HTTP calls and 145 Global API calls, counted at their respective stages of the request pipeline. The 300-step run takes 37.09 hours, including analysis and scoring. In the Commit32 configuration, 19 comparisons require 2,432 additional environment rollouts, whose duration is included in that configuration's step times. Strength reuses existing trajectories but requires additional model-scoring computation.

\section*{Appendix B. Evaluation protocols for published baselines}

\textbf{D2Skill.} Table 1 in the D2Skill paper reports results for Qwen3-4B-Instruct-2507 \cite{d2skill}. We include its SkillRL (O3) reference and D2Skill variants. The default configuration runs for 160 updates, evaluates 128 validation tasks every five updates, and reports the best observed result. The teacher-SFT variant runs for 120 updates.

The default WebShop configuration starts from SFT on SkillRL data. The teacher-SFT variant instead uses 300 successful teacher trajectories per environment, generated by O3 for ALFWorld and Gemini-3-Pro for WebShop. These model revisions and initialization procedures differ from those used in our experiments.

\textbf{ATOD.} The source reports Qwen3-4B validation success after 150 updates for OPD, SOD, TCOD, and ATOD \cite{atod}. Its WebShop column is success rate, and the stated validation batch size is 128. The paper does not specify the complete evaluation panel, ALFWorld split and aggregation, or checkpoint-selection rule.

In the public implementation, success is averaged over trajectories within each batch, then across batches; the calculation does not average the six category rates. This equals the pooled trajectory success rate when there is one batch or all batches have the same size. We place the reported aggregate under ALL in Table 1.

\textbf{ICRL.} The source evaluates Qwen3-4B agents through AgentGym \cite{icrl}. We include its GSPO results, using the ALFWorld and WebShop success columns. WebShop Reward follows the standard terminal matching formula in Appendix D.2 of ICRL; we label it Score and label complete success as Succ. The source defines ALFWorld success as binary task completion and WebShop success as a reward of one, with a thirty-turn limit. It does not explicitly give the evaluation-panel size, split, or checkpoint-selection rule.

The public implementation deduplicates episodes, retains each episode's final executor, and divides successful episodes by the total number of episodes. This is an overall success rate rather than an unweighted category average, so we report it under ALL in Table 1.

Underlining in Table 1 distinguishes published values from our reproduced results. Unreported ATOD and ICRL category results, and ATOD's WebShop Score, are shown as dashes. These published values provide context; the differences attributed to our experiments are computed between our reproduced methods. The source inventory also records additional published results and exclusions arising from different model sizes or ALFWorld splits.

\section*{Appendix C. Additional analyses of response-based skill selection}

\subsubsection*{Do OPD gains require an intact skill structure?}

We compare OPD training with intact skills, sentence-shuffled skills, and neutral text, each containing 2,355 words. Neutral text is used in the OPD branch rather than serving only as an inference-time baseline. Across the four shared validation steps, 90, 105, 120, and 135, intact skills exceed neutral text by 18.4 percentage points, shuffled skills exceed it by 14.3 points, and intact skills exceed shuffled skills by 4.1 points.

Training still benefits after the sentences are shuffled, so an intact procedural structure is not the sole explanation for the OPD gains. This observation motivates examining how text changes model responses. Whether a Local skill is worth retaining is a separate question, tested through the selection experiments in Sections~\ref{sec:global_selection} and~\ref{sec:selection_criteria}.

\subsubsection*{Historical candidate pool and score updates}

The metric comparison in Section~\ref{sec:selection_criteria} compares the three response-based criteria in Equation (3) with Random for long-term skill retention. This earlier configuration accumulates candidates by task category, deduplicates whitespace-normalized text, and keeps at most eight candidates per category. Valid scores are recorded before ranking and capacity-based eviction. The top 50\% of the remaining pool enters Global synthesis, so a criterion affects both pool membership and consolidation. The main framework instead uses independent recent windows and clears all candidates and scores after each window.

If a pooled candidate is scored repeatedly, its first valid score initializes the metric. Later scores are combined by an exponential moving average with coefficient .1: .9 on the previous value and .1 on the new value. Contrast updates its contrastive score only when a scoring event contains both successful and failed trajectories; until the first valid contrastive score, it falls back to Utility.

Random uses a fixed priority obtained by hashing ranking seed zero with the normalized skill text. Candidates without valid scores cannot enter Global selection, and an empty selection leaves Global unchanged. In the main framework, selection reads the Strength score cached at skill generation rather than re-scoring candidates at window boundaries.

Normalized AUC is computed from ten validation success rates at steps 30, 60, ..., 300. We sum the trapezoidal areas between adjacent measurements and divide by the 270-step interval, then express the result as a percentage. AUC therefore summarizes mean validation performance over steps 30--300, while the peak reports the highest observed success rate.

\subsubsection*{Do training gains imply an immediate benefit from skill prompts?}

A separate paired probe holds the policy parameters fixed and compares late-stage execution with and without injected text. Injection-minus-baseline differences are -3.53 percentage points for neutral text, -19.18 for shuffled text, -24.00 for the tested intact text, and -29.24 for antonym replacement. These texts come from a historical pool distinct from the fixed-guidance experiment in Section 3.1.

For this checkpoint and text configuration, direct injection does not improve task execution. The effect of teacher-side guidance during training must therefore be evaluated separately from the effect of adding that text to the policy's execution context. Response-based selection should be used with synthesis that preserves applicability conditions and with subsequent task evaluation.

\subsubsection*{Early Global-injection probes}

An earlier ALFWorld configuration, labeled ALFWorld Full in the records, contains 79 valid ON/OFF probes with the same model parameters and training tasks. A Clean probe after update 110 uses eight tasks and two rollouts per condition: 14/16 succeed with Global ON and 5/16 with Global OFF. This provides an example of immediate skill-guidance gains for that policy and task set.

These probes are supplementary and are separate from the main SkillDuet checkpoint evaluation. The available training logs indicate that the injected trajectories were excluded from parameter updates in this configuration. Complete deployment and task-level probe records are unavailable.

\section*{Appendix D. Retention ratios and optional Global deployment checks}

Table 4 gives the numerical results for Figure~\ref{fig:selection}. Table 5 and Figure~\ref{fig:commit} evaluate the optional Commit check. Appendix A specifies the evaluation used for each comparison.

\par\medskip\noindent\begin{minipage}{\textwidth}\small\setlength{\tabcolsep}{4pt}\renewcommand{\arraystretch}{1.15}
\noindent\textbf{Table 4.} Retention-ratio ablation with recent candidate windows and direct Global updates. Values are peak validation success rates (\%).\par\smallskip
\begin{tabular}{@{}p{0.42\dimexpr\textwidth-16pt\relax}p{0.29\dimexpr\textwidth-16pt\relax}p{0.29\dimexpr\textwidth-16pt\relax}@{}}
\toprule
\textbf{Retention ratio} & \textbf{ALFWorld $\uparrow$} & \textbf{WebShop $\uparrow$} \\
\midrule
Top25\% & \textbf{97.14} & 85.94 \\
\textbf{Top50\%} & \textbf{97.14} & \textbf{92.97} \\
Top75\% & 93.57 & 82.81 \\
\bottomrule\end{tabular}
\end{minipage}\par\medskip

\subsubsection*{Optional Commit checks}

By default, SkillDuet deploys an updated Global after the synthesis checks pass. We also test Commit, an optional acceptance check based on task execution. With policy parameters fixed, Commit compares the candidate and current Global on the same training tasks. Commit32 uses 32 training tasks per Global and two rollouts per task under each condition. These tasks are separate from formal validation. For the comparison tasks $Q$, the mean reward difference is:

\[\begin{aligned}
\Delta_{\mathrm{Commit}}&=\frac{1}{|Q|}\sum_{q\in Q}\Big[\bar R(q,G_{\mathrm{new}})\\
&\qquad-\bar R(q,G_{\mathrm{old}})\Big].
\end{aligned}\tag{4}\]

The candidate replaces the current Global only when all 32 valid task comparisons are available and the mean reward difference is nonnegative. Otherwise, the current skill set is retained. This rule does not use a significance test. Commit controls which Global is deployed; it does not roll back policy updates already made using Local skills.

\par\medskip\noindent\begin{minipage}{\textwidth}\small\setlength{\tabcolsep}{4pt}\renewcommand{\arraystretch}{1.15}
\noindent\textbf{Table 5.} Direct Global updates versus Commit-based acceptance. Values are peak validation success rates (\%).\par\smallskip
\begin{tabular}{@{}p{0.6\dimexpr\textwidth-16pt\relax}p{0.2\dimexpr\textwidth-16pt\relax}p{0.2\dimexpr\textwidth-16pt\relax}@{}}
\toprule
\textbf{Deployment setting} & \textbf{ALFWorld $\uparrow$} & \textbf{WebShop $\uparrow$} \\
\midrule
\textbf{Top50\%, direct update} & \textbf{97.14} & \textbf{92.97} \\
Top25\% + Commit32 & 92.14 & 89.06 \\
Top50\% + Commit32 & 96.43 & 85.94 \\
Top75\% + Commit32 & 95.00 & 89.84 \\
\bottomrule\end{tabular}
\end{minipage}\par\medskip

At 50\% retention, direct updates achieve peak validation success rates of 97.14\% on ALFWorld and 92.97\% on WebShop, compared with 96.43\% and 85.94\% with Commit32. We therefore use direct updates in the main configuration. Table 5 and Figure~\ref{fig:commit} also include Commit32 at the other retention ratios.

\par\medskip\noindent\begin{minipage}{\textwidth}\centering\includegraphics[width=\textwidth]{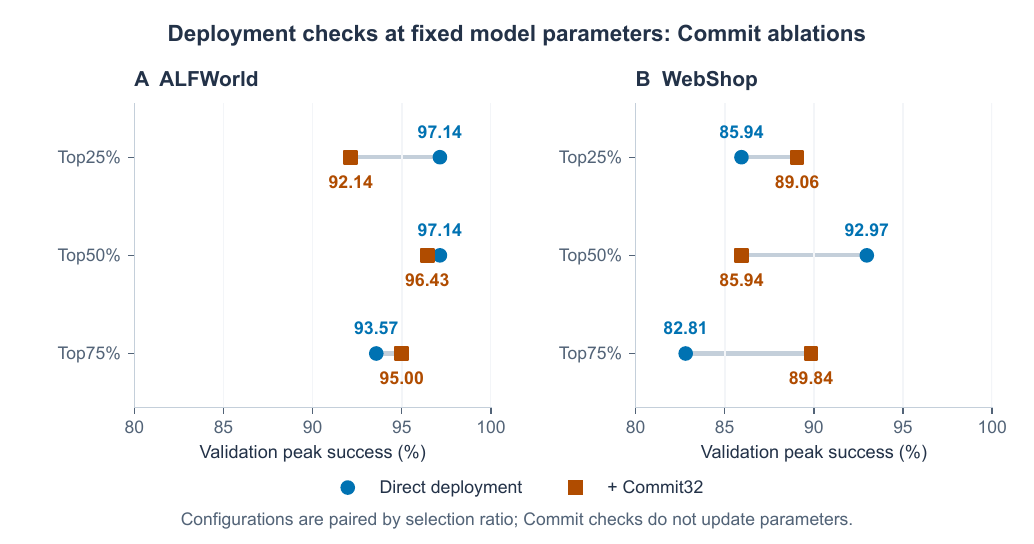}\par\smallskip\small\refstepcounter{figure}\noindent\textbf{Figure \thefigure}\enspace|\enspace Optional execution-based acceptance checks for Global. Direct deployment and Commit32 are compared at matched retention ratios. At Top50\% retention, direct deployment outperforms Commit32 on both benchmarks. Commit controls Global deployment without reversing Local-guided policy updates.\label{fig:commit}\end{minipage}\par\medskip

\section*{Appendix E. Detailed results at the selected checkpoints}

For ALFWorld, we first compute each category's success rate by pooling its evaluation trajectories, then take the unweighted mean of the six rates to obtain Avg. ALL is the fraction of successful trajectories across the full evaluation set. Every configuration uses Qwen3-4B and eval\_in\_distribution. The SkillDuet and SEED-API (.05) results in Table 1 come from separate evaluations of their step 300 checkpoints. For the other reproduced ALFWorld methods, we select checkpoints by validation ALL within the training budget, breaking ties in favor of the later checkpoint. All metrics in a row come from the same evaluation; N is the number of trajectories.

\par\medskip\noindent\begin{minipage}{\textwidth}\small\setlength{\tabcolsep}{4pt}\renewcommand{\arraystretch}{1.15}
\noindent\textbf{Table 6.} ALFWorld success rates by category, unweighted category mean (Avg), and overall success (ALL), in percent. Dashes indicate unreported category results.\par\smallskip
\footnotesize\setlength{\tabcolsep}{2pt}
\begin{tabular}{@{}p{0.235\dimexpr\textwidth-40pt\relax}>{\raggedleft\arraybackslash}p{0.065\dimexpr\textwidth-40pt\relax}>{\raggedleft\arraybackslash}p{0.07777777777777778\dimexpr\textwidth-40pt\relax}>{\raggedleft\arraybackslash}p{0.07777777777777778\dimexpr\textwidth-40pt\relax}>{\raggedleft\arraybackslash}p{0.07777777777777778\dimexpr\textwidth-40pt\relax}>{\raggedleft\arraybackslash}p{0.07777777777777778\dimexpr\textwidth-40pt\relax}>{\raggedleft\arraybackslash}p{0.07777777777777778\dimexpr\textwidth-40pt\relax}>{\raggedleft\arraybackslash}p{0.07777777777777778\dimexpr\textwidth-40pt\relax}>{\raggedleft\arraybackslash}p{0.07777777777777778\dimexpr\textwidth-40pt\relax}>{\raggedleft\arraybackslash}p{0.07777777777777778\dimexpr\textwidth-40pt\relax}>{\raggedleft\arraybackslash}p{0.07777777777777778\dimexpr\textwidth-40pt\relax}@{}}
\toprule
\textbf{Method (OPD coefficient)} & \textbf{step} & \textbf{Pick $\uparrow$} & \textbf{Look $\uparrow$} & \textbf{Clean $\uparrow$} & \textbf{Heat $\uparrow$} & \textbf{Cool $\uparrow$} & \textbf{Pick2 $\uparrow$} & \textbf{Avg $\uparrow$} & \textbf{ALL $\uparrow$} & \textbf{N} \\
\midrule
Vanilla & 0 & 70.83 & 40.00 & 19.75 & 18.52 & 20.00 & 26.67 & 32.63 & 37.38 & 420 \\
SEED (.01) & 135 & 100.00 & 100.00 & 100.00 & 100.00 & 90.48 & 87.50 & 96.33 & 96.09 & 128 \\
SEED-API (.01) & 240 & --- & --- & --- & --- & --- & --- & --- & $\approx$88.6 & 140 \\
SEED-API (.05) & 300 & 100.00 & 93.33 & 96.30 & 83.33 & 95.00 & 100.00 & 94.66 & 95.71 & 140 \\
\textbf{SkillDuet Top50\% (.05)} & 300 & 100.00 & 100.00 & 100.00 & 100.00 & 90.00 & 100.00 & \textbf{98.33} & \textbf{98.57} & 140 \\
GRPO & 300 & 100.00 & 85.71 & 94.29 & 61.54 & 82.76 & 72.22 & 82.75 & 86.43 & 140 \\
SDAR (.01) & 270 & 100.00 & 100.00 & 100.00 & 54.55 & 70.37 & 80.95 & 84.31 & 87.86 & 140 \\
Earlier SkillDuet + Commit16 (.05) & 300 & 100.00 & 75.00 & 100.00 & 94.12 & 86.96 & 100.00 & 92.68 & 95.00 & 140 \\
Earlier SkillDuet + Commit32 (.03) & 300 & 100.00 & 88.89 & 100.00 & 100.00 & 85.19 & 95.00 & 94.85 & 95.71 & 140 \\
\bottomrule\end{tabular}
\end{minipage}\par\medskip

SEED-API (.05) and SkillDuet succeed on 134/140 and 138/140 trajectories, respectively, using identical category denominators of 40/15/27/18/20/20. SkillDuet improves Avg by 3.67 percentage points and ALL by 2.86 points.

SEED, GRPO, and SDAR succeed on 123/128, 121/140, and 123/140 trajectories, with category denominators of 36/11/26/10/21/24, 31/14/35/13/29/18, and 29/15/37/11/27/21, respectively. The two earlier Commit configurations yield 133/140 and 134/140 successes; they are distinct from the retention-matched configurations in Table 5.

WebShop Score is the mean continuous terminal reward, and Succ. is the complete-task success rate. Table 7 reports the checkpoint with the highest validation Succ., with both Score and Succ. taken from the same evaluation, consistent with the validation-peak ablations. All configurations use Qwen3-4B, catalog1000, and the same task split and sampling rules. An evaluation contains 128 executions, which may include repeated goals. Where the complete goal-ID sequence is unavailable, we compare aggregate success rates rather than paired outcomes for individual tasks.

\par\medskip\noindent\begin{minipage}{\textwidth}\small\setlength{\tabcolsep}{4pt}\renewcommand{\arraystretch}{1.15}
\noindent\textbf{Table 7.} WebShop Score and Succ. (\%) at checkpoints selected by validation Succ.\par\smallskip
\begin{tabular}{@{}p{0.6\dimexpr\textwidth-24pt\relax}>{\raggedleft\arraybackslash}p{0.1\dimexpr\textwidth-24pt\relax}>{\raggedleft\arraybackslash}p{0.15\dimexpr\textwidth-24pt\relax}>{\raggedleft\arraybackslash}p{0.15\dimexpr\textwidth-24pt\relax}@{}}
\toprule
\textbf{Method} & \textbf{step} & \textbf{Score $\uparrow$} & \textbf{Succ. $\uparrow$} \\
\midrule
Vanilla & 0 & 13.85 & 0.00 \\
SEED & 150 & 86.23 & 75.78 \\
SEED-API (.03) & 270 & 91.31 & 87.50 \\
GRPO & 270 & 92.96 & 89.84 \\
SDAR (.01) & 270 & 89.83 & 86.72 \\
\textbf{SkillDuet Top50\% (.03)} & 270 & \textbf{95.18} & \textbf{92.97} \\
Paired Local-only (.03) & 270 & 92.53 & 87.50 \\
SkillDuet Random Top50\% (.03) & 270 & 92.78 & 89.06 \\
SkillDuet + Commit32 (.03) & 180 & 93.18 & 85.94 \\
\bottomrule\end{tabular}
\end{minipage}\par\medskip

\textbf{Unadapted Qwen3-4B.} Vanilla receives no task-specific SFT or RL and uses no Local skills, Global skills, or external analyzer. Its ALFWorld evaluation uses the same task panel as the controlled Global study, repeated under three evaluation seeds for 420 episodes in total; Table 6 reports this evaluation size. Its WebShop evaluation uses 128 trajectories covering 111 distinct catalog1000 goals with one seed. Score and Succ. in Table 7 come from that same evaluation.

\small

\section*{Appendix F. Controlled evaluations of Global skills}

\subsubsection*{Model checkpoints and text conditions}

SkillDuet and SEED-API use their Qwen3-4B checkpoints from step 300, trained with an OPD coefficient of .05. SEED-API receives no task-specific SFT, generates Local skills with the external analyzer, and trains without Global. All conditions share 140 valid\_seen episode slots covering 99 distinct games, with identical initial states. Evaluation seeds 0/1/2 give 420 executions per condition. Both conditions with learned Global use the SkillDuet skill set deployed at step 291; transfer does not change SEED-API's parameters.

Decoding uses temperature=.4, top\_p=1, top\_k=-1, input and response limits of 2,048 and 512 tokens, five history turns, and at most thirty actions.

Neutral replacement preserves Global's prompt position and format. At each state, the complete prompt containing neutral text is matched in token length to the complete prompt containing learned Global, without truncation. All 6,714 audited active requests satisfy exact token-length matching. Vanilla is evaluated on the same task panel without task-specific SFT/RL or Local/Global guidance (Table 8).

Under their usual deployment conditions, SkillDuet with Global exceeds SEED-API without Global by 3.81 percentage points (Table 8).

\par\medskip\noindent\begin{minipage}{\textwidth}\small\setlength{\tabcolsep}{4pt}\renewcommand{\arraystretch}{1.15}
\noindent\textbf{Table 8.} Complete ALFWorld results for the controlled Global evaluations (\%). Each condition pools three evaluation seeds, with category denominators of 120/45/81/54/60/60.\par\smallskip
\begin{tabular}{@{}p{0.16\dimexpr\textwidth-64pt\relax}>{\raggedleft\arraybackslash}p{0.105\dimexpr\textwidth-64pt\relax}>{\raggedleft\arraybackslash}p{0.105\dimexpr\textwidth-64pt\relax}>{\raggedleft\arraybackslash}p{0.105\dimexpr\textwidth-64pt\relax}>{\raggedleft\arraybackslash}p{0.105\dimexpr\textwidth-64pt\relax}>{\raggedleft\arraybackslash}p{0.105\dimexpr\textwidth-64pt\relax}>{\raggedleft\arraybackslash}p{0.105\dimexpr\textwidth-64pt\relax}>{\raggedleft\arraybackslash}p{0.105\dimexpr\textwidth-64pt\relax}>{\raggedleft\arraybackslash}p{0.105\dimexpr\textwidth-64pt\relax}@{}}
\toprule
\textbf{Condition} & \textbf{Pick $\uparrow$} & \textbf{Look $\uparrow$} & \textbf{Clean $\uparrow$} & \textbf{Heat $\uparrow$} & \textbf{Cool $\uparrow$} & \textbf{Pick2 $\uparrow$} & \textbf{Avg $\uparrow$} & \textbf{ALL $\uparrow$} \\
\midrule
Vanilla & 70.83 & 40.00 & 19.75 & 18.52 & 20.00 & 26.67 & 32.63 & 37.38 \\
SEED-API: none & 99.17 & 86.67 & 95.06 & 87.04 & 93.33 & 93.33 & 92.43 & 93.81 \\
SEED-API: learned & 95.83 & 88.89 & 95.06 & 90.74 & 85.00 & 75.00 & 88.42 & 89.76 \\
SkillDuet: none & 99.17 & 75.56 & 55.56 & 53.70 & 68.33 & 90.00 & 73.72 & 76.67 \\
SkillDuet: neutral & 99.17 & 84.44 & 56.79 & 42.59 & 63.33 & 86.67 & 72.17 & 75.24 \\
\textbf{SkillDuet: learned} & 98.33 & \textbf{100.00} & \textbf{100.00} & \textbf{100.00} & \textbf{95.00} & 91.67 & \textbf{97.50} & \textbf{97.62} \\
\bottomrule\end{tabular}
\end{minipage}\par\medskip

\subsubsection*{Paired outcomes and uncertainty estimates}

Learned Global, removal, and neutral replacement yield 410, 322, and 316 successes, respectively. For learned Global versus removal, 315 pairs succeed in both conditions, 95 succeed only with learned Global, seven succeed only without it, and three fail in both. For learned Global versus neutral replacement, 308 pairs succeed in both conditions, 102 succeed only with learned Global, eight succeed only with neutral text, and two fail in both. For neutral replacement versus removal, 284 pairs succeed in both conditions, 32 succeed only with neutral text, 38 succeed only with removal, and 66 fail in both.

We use 10,000 paired bootstrap draws, stratified by task category and clustered by unique game. All repeated slots and evaluation seeds for a game are resampled together. Learned Global improves ALL by 20.95 percentage points over removal, with a 95\% interval of [14.07, 27.96], and by 22.38 points over neutral replacement, with an interval of [15.93, 29.23]. Neutral replacement differs from removal by -1.43 points, with an interval of [-4.57, 1.93]. Adding the same Global to SEED-API changes success by -4.05 points, with an interval of [-8.45, 0.47]. These intervals describe variability from resampling the observed games at fixed model checkpoints.

Compared with neutral replacement, learned Global yields a net gain of 94 successes overall, including 35 on Clean, 31 on Heat, and 19 on Cool. Figure~\ref{fig:global_ablation} gives category-level results for the removal comparison. These results come from the independent fixed-checkpoint evaluation, not the main-table evaluation.

\par\medskip\noindent\begin{minipage}{\textwidth}\centering\includegraphics[width=\textwidth]{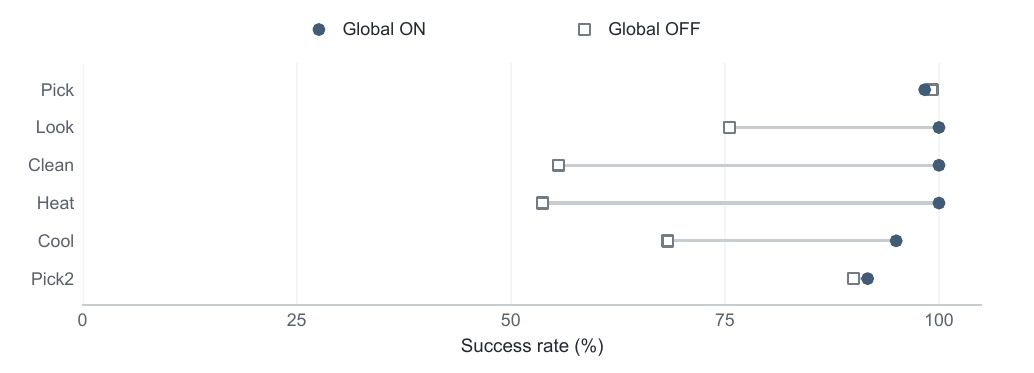}\par\smallskip\small\refstepcounter{figure}\noindent\textbf{Figure \thefigure}\enspace|\enspace Effect of removing Global from SkillDuet on ALFWorld, with model parameters fixed. The largest gains from Global occur on Clean, Heat, and Cool.\label{fig:global_ablation}\end{minipage}\par\medskip

\subsubsection*{Paired action-sequence example}

The Clean example in Section~\ref{sec:global_compatibility} is the first pair in which learned Global succeeds and removal fails, ordered by evaluation seed, batch, and slot (0/0/0; ButterKnife--DiningTable). With Global, the agent follows the clean-before-place procedure and completes the task in 7 actions. Without Global, it reaches the 30-action limit without completing the task.

Full trajectories, prompts, and deployment configurations are available for the 840 episodes with learned Global or its removal. The additional SEED-API, neutral-text, and Vanilla statistics are computed from episode-level completion records. Request-level audits establish neutral-text length matching. All interventions change the execution context while leaving model parameters fixed.

\subsubsection*{Scope of the comparison}

The three evaluation seeds measure execution variability at these checkpoints, not variability across independent training runs. Appendices A, B, and E give the training budgets, evaluation protocols, category results, and denominators used in the main comparisons.

\section*{Appendix G. A worked example of hierarchical Global synthesis}

\textbf{Turning repeated omissions into a reusable procedure.} In this WebShop example, several trajectories omit required product attributes: L0006 omits fit type, L0021 omits round shape, and L0034 omits youth. Their Local skills record these failures. Synthesis combines the recurring lesson into a general rule: explicitly select each required field and avoid repeated clicking (Figure~\ref{fig:case}).

\textbf{Compressing experience while preserving conditions and sources.} Inputs that fit the context budget are synthesized directly. Larger inputs are summarized in batches, retaining shared procedures, applicability conditions, and source identifiers; compression repeats if necessary. Here, the 37 selected Local skills from steps 11--20 produce 14 summaries in four batches, then eight summaries in two batches. These summaries are combined with five existing Global rules to produce five updated rules. Both summary layers retain the sources of the attribute omissions above. The previous G4 rule is retained verbatim, while other rules are revised or combined.

\textbf{Reusing the updated skills.} The new Global is deployed before interaction at step 21 and is used during training. The same skill set is recorded as the prior Global at step 31. Thus, the consolidated experience guides subsequent interactions and is reused as input to the next Global update.

\par\medskip\noindent\begin{minipage}{\textwidth}\centering\includegraphics[width=.94\textwidth]{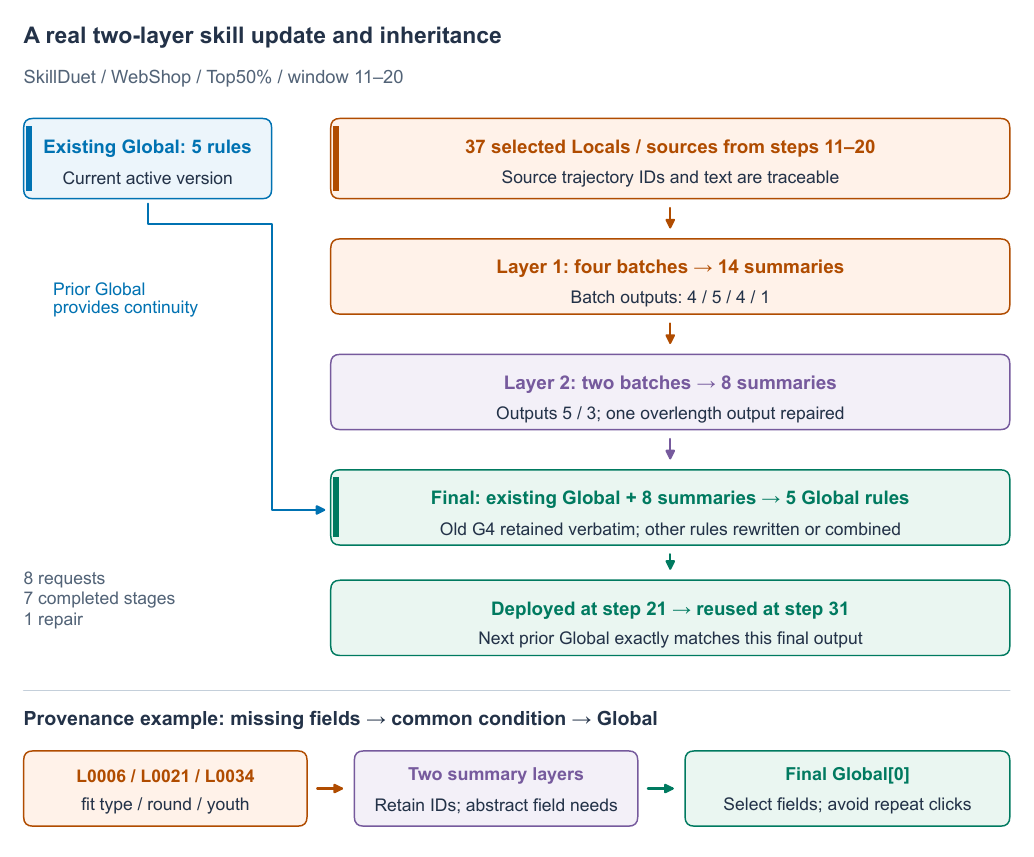}\par\smallskip\small\refstepcounter{figure}\noindent\textbf{Figure \thefigure}\enspace|\enspace Hierarchical skill consolidation on WebShop. Thirty-seven selected Local skills produce fourteen first-layer and eight second-layer summaries, which are combined with five existing Global rules to yield five updated rules. The update uses eight requests, including one repair. Deployment at step 21 and reuse as the prior Global at step 31 are verified in the records. The updated skill set guides subsequent interactions and further consolidation. Skill excerpts are paraphrased.\label{fig:case}\end{minipage}\par\medskip
\end{document}